\documentclass[runningheads]{llncs}
\usepackage{booktabs}
\usepackage{multirow}
\usepackage{amsmath}
\usepackage{amssymb}
\usepackage{amsfonts}
\usepackage{makecell}
\usepackage[T1]{fontenc}
\usepackage{bm}
\usepackage{cite}
\usepackage{array}
\usepackage{url}
\usepackage[misc]{ifsym}
\usepackage{bbding}
\usepackage[colorlinks,
            linkcolor=blue,
            anchorcolor=blue,
            citecolor=blue]{hyperref}

\usepackage{graphicx}
\begin{document}
\title{FiRework: Field Refinement Framework for Efficient Enhancement of Deformable Registration}
\titlerunning{FiRework: Field Refinement Framework for Registration}
%
\author{Haiqiao Wang \and Dong Ni \and Yi Wang\Envelope}
\authorrunning{H. Wang et al.}
%
\institute{Smart Medical Imaging, Learning and Engineering (SMILE) Lab,\\
	Medical UltraSound Image Computing (MUSIC) Lab,\\
	School of Biomedical Engineering,
	Shenzhen University Medical School,\\
	Shenzhen University, Shenzhen, China\\
	\email{onewang@szu.edu.cn} \\}
%
\maketitle              
\begin{abstract}
Deformable image registration remains a fundamental task in clinical practice, yet solving registration problems involving complex deformations remains challenging.
Current deep learning-based registration methods employ continuous deformation to model large deformations, which often suffer from accumulated registration errors and interpolation inaccuracies.
Moreover, achieving satisfactory results with these frameworks typically requires a large number of cascade stages, demanding substantial computational resources.
Therefore, we propose a novel approach, the field refinement framework (FiRework), tailored for unsupervised deformable registration, aiming to address these challenges.
In FiRework, we redesign the continuous deformation framework to mitigate the aforementioned errors.
Notably, our FiRework requires only one level of recursion during training and supports continuous inference, offering improved efficacy compared to continuous deformation frameworks.
We conducted experiments on two brain MRI datasets, enhancing two existing deformable registration networks with FiRework.
The experimental results demonstrate the superior performance of our proposed framework in deformable registration.
\textit{The code is publicly available at \url{https://github.com/ZAX130/FiRework}}.

\end{abstract}
\section{Introduction}
Deformable image registration is a crucial foundational task in clinical practice, providing aligned information to assist doctors in diagnosis and interventions.
The purpose of registration is to estimate the desired deformation field that warps the moving image to align with the fixed image.
While numerous methods have been proposed to address deformable registration, including traditional approaches~\cite{Bajcsy1989, Rueckert1999, LDDMM, AVANTS2008}  and deep learning methods~\cite{VoxelMorph, Chen2022a}, tackling complex deformations remains challenging.

\begin{figure}[t]
	\centering
	\includegraphics[width=0.99\columnwidth]{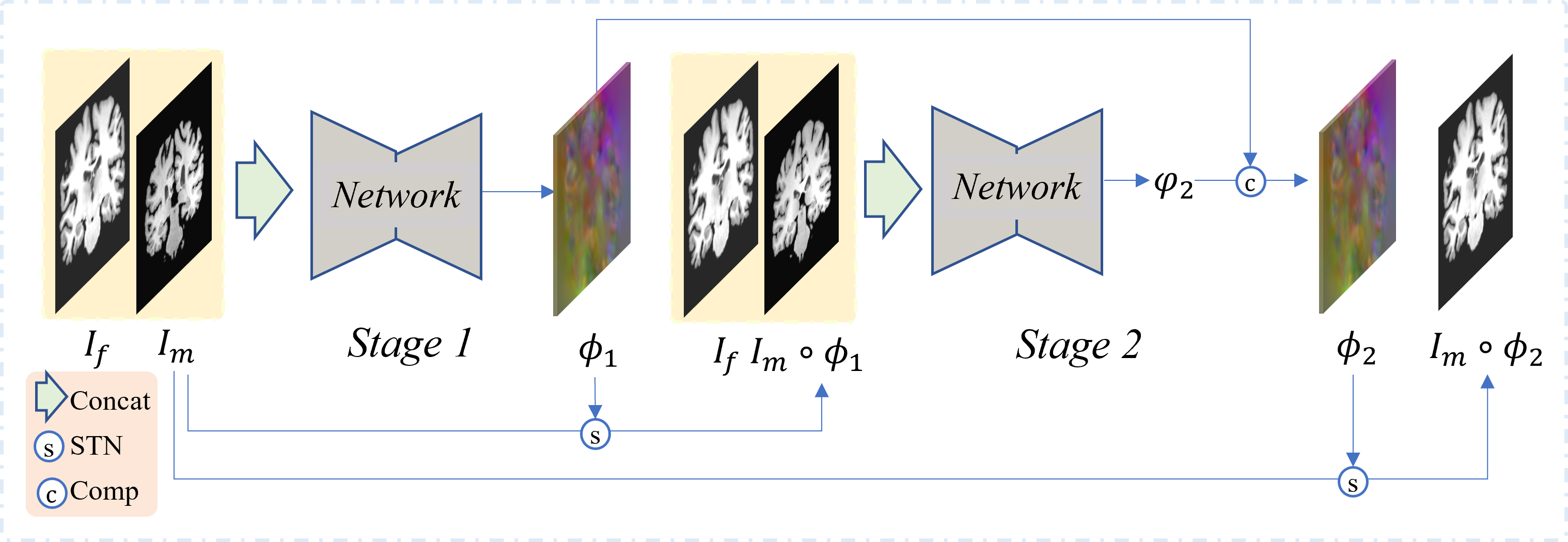}
	\caption{Illustration of a two-stage continuous registration framework. $I_f$ and $I_m$ are the fixed and moving images, respectively. The network at stage 1 predicts the initial deformation field $\phi_1$. Then the warped moving image $I_m \circ \phi_1$, together with $I_f$, are sent into the network at stage 2 to compute the residual deformation field $\varphi_2$. Finally, $\phi_1$ and $\varphi_2$ are fused together to generate the overall deformation field $\phi_2$ for registering.}
	\label{f:VM-2}
\end{figure}

To address this issue, recent studies relying on continuous deformation have been proposed~\cite{Zhao2019, zheng2022recursive, Dual, RDN, Zhu2021, Wang2023, wang2024recursive, wang2024modetv2}.
These structures typically involve computing intermediate deformation fields to warp images or feature maps, followed by further residual field computation, and finally aggregating all fields to obtain the overall deformation field (see Fig.~\ref{f:VM-2} for illustration).
According to~\cite{Zhao2019}, the formula for obtaining $\phi_t$ from the deformation field $\phi_{t-1}$ is as follows:
\begin{equation}
	\begin{split}
		\varphi_{t} &= f(I_m \circ \phi_{t-1}, I_f), \\
		\phi_{t} &= comp( \phi_{t-1}, \varphi_{t}),
		\label{e:continue}
	\end{split}
\end{equation}
where $\varphi_t$ represents the residual subfield at step $t$, $\circ$ represents the warping operation,
typically performed by the spatial transformer network (STN)~\cite{NIPS2015_33ceb07b},
\( f \) represents either a cascaded or recurrent network, with \( I_m \) and \( I_f \) being the moving and fixed images, respectively, and $comp$ means the composition of continuous deformation field.
If \( f \) in Equation~(\ref{e:continue}) is a pyramid model, \( I_m \) and \( I_f \) represent the feature maps of a certain level for the moving and fixed images. 
However, such structures often encounter certain issues, such as the unavoidable impact of interpolation errors on feature extraction, as well as the influence of cumulative errors on the final deformation performance~\cite{Zhu2021}. 
Moreover, multi-stage deformation structures often require cascading multiple deformation field generation networks during training~\cite{Zhao2019}, leading to increased network complexity and reduced training efficiency.

We have reconsidered the two types of error above. Firstly, we contend that relying solely on image information is insufficient to overcome the local minima induced by accumulated errors. This is primarily because the network lacks the knowledge of how \( I_m \) transforms to \( I_m\circ\phi_{t-1} \) at stage \( t \), thus unable to consider or address the origin of such errors, resulting in the forced convergence of \( I_m\circ\phi_{t-1} \) towards \( I_f \). Secondly, while interpolation errors might seem addressable by directly inputting \( I_m \), the objective of network is to compute  \( \varphi_t \) on the basis of \( I_m\circ\phi_{t-1} \), rendering the utility of \( I_m \) incomprehensible.

Recent attempts to address this problem by propagating semantic feature maps~\cite{meng2023non, wang2024recursive} or re-weighting multi-level deformation fields~\cite{Lv2022, Chen2022}. However, these approaches often lead to gradient explosion or substantial computational resource consumption.

In light of the challenges, we propose a novel field refinement framework (FiRework) for deformable registration networks.
Unlike existing continuous deformation frameworks, our framework treats the network as a deformable field refiner.
In this framework, we directly incorporate \( I_m \) and \( \phi_{t-1} \), as the relationship between \( I_m \) and \( I_m \circ \phi_{t-1} \),  into the network as extra inputs, allowing the network to reevaluate and optimize the previous deformation at each stage, thereby mitigating the issue of accumulated errors, and can leverage \( I_m \) to address interpolation errors. 
Moreover, our framework requires only a single-level recursion during training and allows for continuous refinement of the deformation field during inference, effectively reducing training costs and improving registration performance.
The main contributions of our work are summarized as follows:
\begin{itemize}
	\item[$\bullet$] We propose a novel deformable registration framework that bypasses interpolation and accumulation errors associated with continuous deformation by performing continuous deformation field refinement.
	
	\item[$\bullet$] The proposed framework entails a single-level recursion during training, while enabling multiple iterations for deformation field refinement during inference.

	\item[$\bullet$] We adapt two existing registration networks to the proposed framework, demonstrating substantial performance improvements in experimental results.
\end{itemize}

\section{Method}
\subsection{Training Process}
\begin{figure}[t]
	\centering
	\includegraphics[width=0.99\columnwidth]{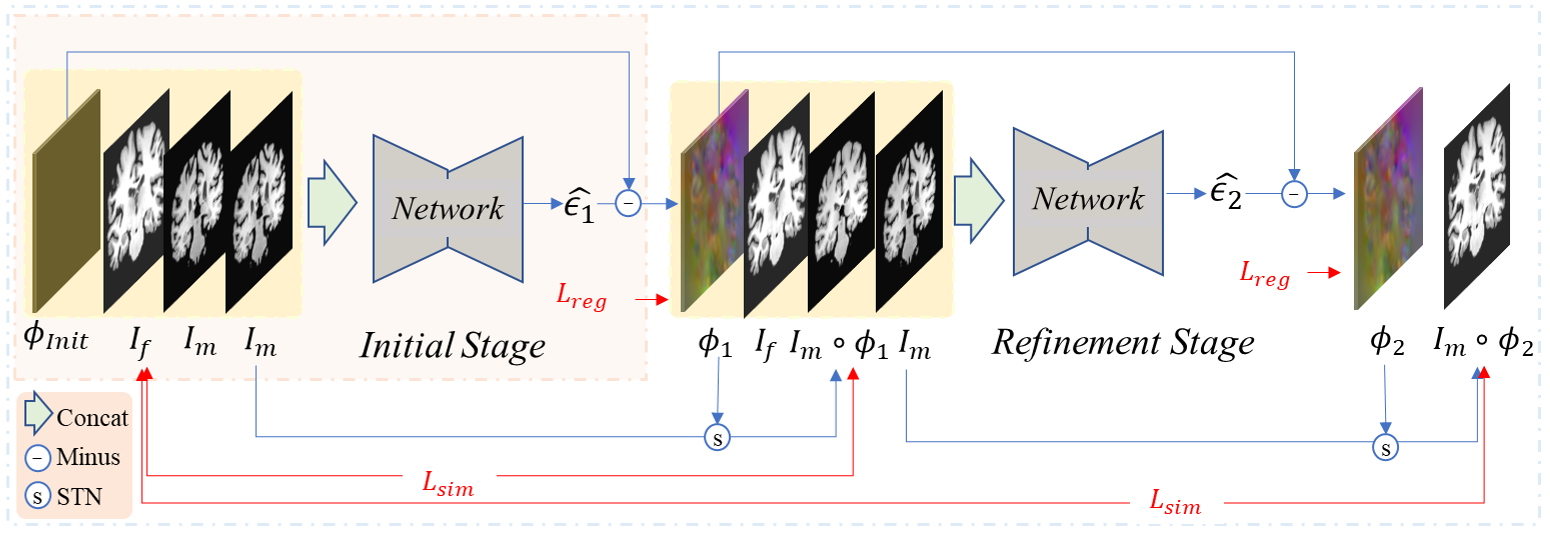}
	\caption{Illustration of the training process of the proposed field refinement framework (FiRework). The training process is primarily divided into the initial stage and refinement stage.}
	\label{f:Training}
\end{figure}

Our overall training process consists of two stages:
the initialization stage and refinement stage, as illustrated in Fig.~\ref{f:Training}.
Given the fixed image \( I_f \) and the moving image \( I_m \), in the initialization stage, the network is modeled as follows:
\begin{equation}
	\hat{\epsilon_1}=f(I_m, I_m, I_f, \phi_{Init}),
	\label{e:p1}
\end{equation}
where \( f \) denotes the network, \( \phi_{Init} \) represents the initial deformation field initialized to zero, and $\hat{\epsilon_1}$ signifies the errors in the input deformation field.
Thus, from the initialization stage, we obtain the first-stage deformation field \( \phi_1 \) as:
\begin{equation}
	\phi_1=\phi_{Init}-\hat{\epsilon_1}=-\hat{\epsilon_1}.
	\label{e:phi1}
\end{equation}
It is noteworthy that Equation~(\ref{e:p1}) signifies that in the initialization stage, the network primarily relies on \( I_m \) and \( I_f \) to solve for the deformation field, which is consistent with the modeling of single-stage registration networks.
We repeatedly input \( I_m \) in the initialization stage, and the input of the initial deformation field is primarily for consistency with the subsequent refinement stage.
Through the initialization stage, we obtain a rough deformation field generated by the network.

To further improve the quality of the deformation field, we introduce a refinement stage. In the refinement stage, the network is modeled as follows:
\begin{equation}
	\begin{split}
	\hat{\epsilon_2} = &f(I_m, I_m \circ \phi_1, I_f, \phi_1), \\
&\phi_2 = \phi_1-\hat{\epsilon_2},
\label{e:phi2}
	\end{split}
\end{equation}
where $\circ$ is the warping operation and \( I_m \circ \phi_1 \) represents the image obtained by deforming the moving image using the first-stage deformation field $\phi_1$.
During this process, the network learns the errors $\hat{\epsilon_2}$ in \( \phi_1 \) based on the conditions \( I_m, I_m \circ \phi_1, I_f, \) and \( \phi_1 \), and subsequently eliminates it from \( \phi_1 \) to derive the refined deformation field \( \phi_2 \).
We design it to output $\hat{\epsilon}$ instead of directly providing the optimized deformation field in order to alleviate the workload of the network and enable it to utilize the input deformation field. 

Because the network at stage \( t \) directly obtains refinement on \( \phi_{t-1} \), rather than the continuous deformation as in Equation~(\ref{e:continue}), \( I_m \circ \phi_{t-1} \) serves only as a reference, and the network focuses on learning how to better utilize the refined deformation field \( \phi_{t} \) to directly deform \( I_m \) to \( I_f \).
Consequently, our proposed approach circumvents interpolation and accumulation errors inherent in continuous deformation frameworks.

To guide the learning process of the FiRework, our loss function is the sum of four losses as depicted in Fig.~\ref{f:Training}:
\begin{equation}
	\mathcal{L}=\mathcal{L}_{\text{sim}}(I_m \circ \phi_1, I_f)+\lambda\mathcal{L}_{\text{reg}}(\phi_1)+\mathcal{L}_{\text{sim}}(I_m \circ \phi_2, I_f)+\lambda\mathcal{L}_{\text{reg}}(\phi_2),
\end{equation}
where \( \mathcal{L}_{\text{sim}} \) represents the normalized cross correlation loss~\cite{rao2014application},
\(\mathcal{L}_{\text{reg}} \) denotes the square of the gradient of the deformation field~\cite{VoxelMorph},
and $\lambda$ is the weight of the regularization term.
It is important to note that the network in our framework shares parameters.

\begin{figure}[t]
	\centering
	\includegraphics[width=0.99\columnwidth]{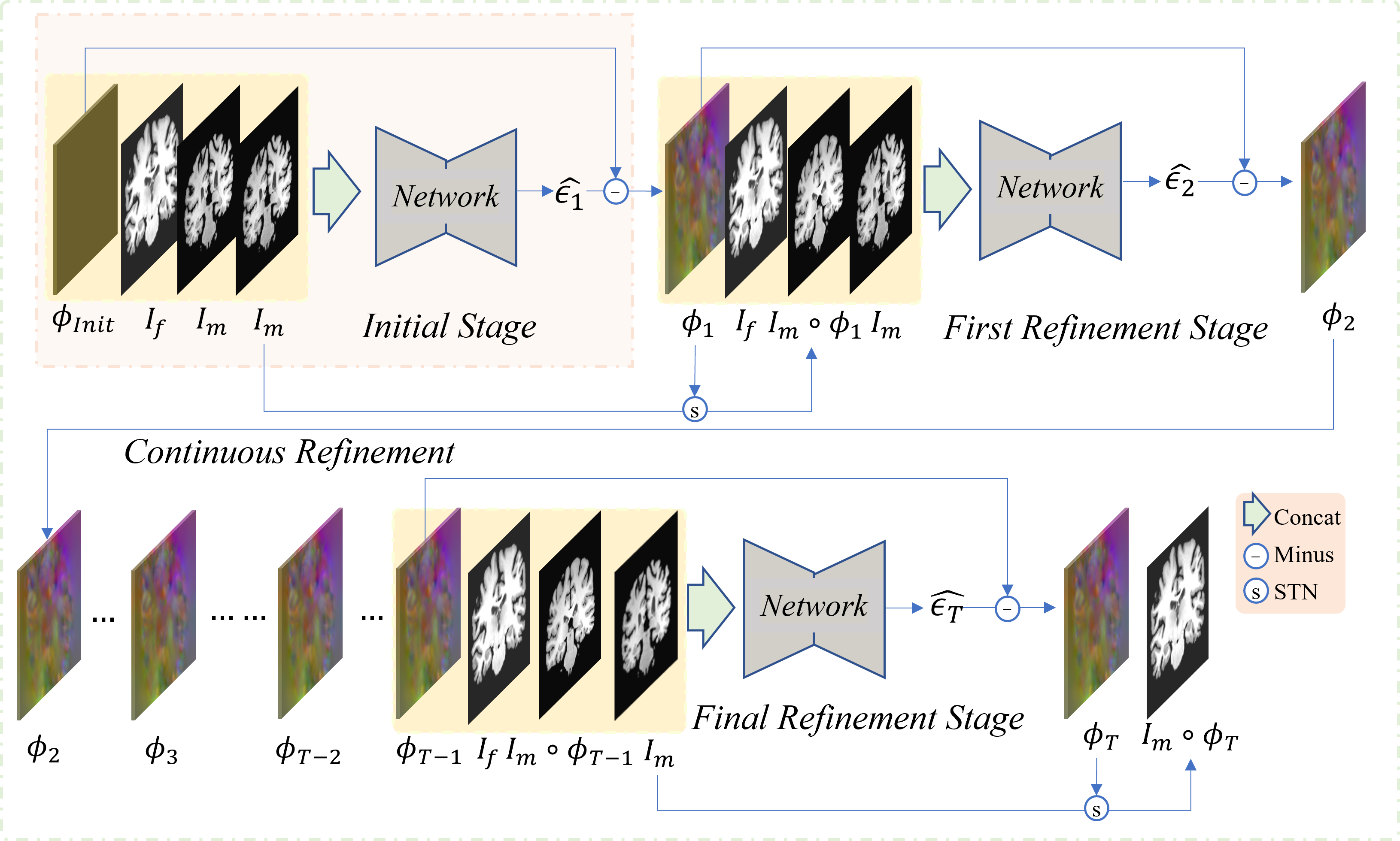}
	\caption{Illustration of the inference process of the proposed FiRework. The figure illustrates the generation process from \( \phi_1 \) to \( \phi_T \).}
	\label{f:Inference}
\end{figure}

\subsection{Inference Process}
After training the network according to the aforementioned training process, we can obtain a network to directly optimize the coarse deformation field $\phi_t$ to obtain the refined deformation field $\phi_{t+1}$, as illustrated in Fig.~\ref{f:Inference}.
During the initialization phase, we employ the same strategy as in the training process.
Initially, we input \(I_m, I_m, I_f, \phi_{\text{Init}}\) into the network to obtain \(\hat{\epsilon}_1\).
Subsequently, we input \(I_m, I_m\circ\phi_1, I_f, \phi_1\) to acquire \(\hat{\epsilon}_2\), and iterate this process iteratively to obtain \(\hat{\epsilon}_3, ..., \hat{\epsilon}_T\).
The final deformation field \(\phi\) is then computed as:
\begin{equation}
	\phi = \phi_T =-\sum_{t=1}^{T} \hat{\epsilon}_t.
\end{equation}
Once the deformation field \(\phi\) is acquired, the moving image $I_m$ can be warped to obtain the registration result $I_m \circ \phi$.

\section{Experiments}
\subsubsection{Datasets.}
Our experiments utilized two publicly accessible brain MRI datasets:
LPBA~\cite{Shattuck2008} and Mindboggle~\cite{Klein2012}.
The LPBA dataset comprises MRI volumes with 54 manually annotated regions of interest (ROIs).
For training, we utilized 30 volumes ($30\times29$ pairs), reserving 10 volumes ($10\times9$ pairs) for testing.
The Mindboggle dataset includes MRI volumes with 62 manually delineated ROIs.
We combined the NKI-RS-22 and NKI-TRT-20 subsets to form the training set, comprising a total of 42 examples ($42\times 41$ pairs), and used 20 volumes from OASIS-TRT-20 ($20\times19$ pairs) for testing. 
Pre-processing involved normalizing all voxel values to the range 0 to 1, and skull-stripping using FreeSurfer~\cite{Fischl2012}, resulting in volumes standardized to a final size of $160\times192\times160$ after center-cropping.

\subsubsection{Evaluation Metrics.}
Quantitative assessment was conducted using Dice score (DSC)~\cite{Dice1945} as the primary metric, measuring the degree of overlap between corresponding regions.
Additionally, the average symmetric surface distance (ASSD) ~\cite{Taha2015} was computed to evaluate the similarity of surfaces.
The quality of the predicted deformation field $\phi$ was evaluated by determining the percentage of voxels with a non-positive Jacobian determinant (i.e., folding ratio).
All metrics were calculated in 3D space, with superior registration indicated by higher DSC, and smaller ASSD and folding ratio.

\subsubsection{Comparison Methods.}
Our FiRework was compared against several state-of-the-art registration techniques:
(1)~\texttt{SyN}~\cite{AVANTS2008}: a classical traditional approach.
(2)~\texttt{XMorpher(XM)}~\cite{shi2022xmorpher}: a registration network incorporating cross-attention Transformer modules for each level of encoder and decoder.
(3)~\texttt{VoxelMorph(VM)}~\cite{VoxelMorph}: a classic one-stage registration network.
(4)~\texttt{TransMorph(TM)}~\cite{Chen2022a}: a single-stage registration network with SwinTransformer enhanced encoder.
(5)~\texttt{DMR}~\cite{Chen2022}: a registration network using a Deformer and a multi-resolution refinement module.
(6)~\texttt{PR++}~\cite{Dual}: a pyramid registration network using 3D correlation layer.

\subsubsection{Implementation Details.}
Our methods included enhancing VM and TM with our FiRework, named FiRework-VM and FiRework-TM.
The Adam optimizer~\cite{kingma2014adam} with a learning rate decay strategy was employed:
\begin{equation}
	lr_m =  lr_{init}\cdot(1-\frac{m-1}{M})^{0.9}, m = 1, 2, ... ,M,
\end{equation}
where $lr_m$ represents the learning rate of $m$-th epoch and $lr_{init}$ represents the learning rate of initial epoch.
For LPBA, we employed $lr_{init} =0.0004$ and $\lambda=4$.
For Mindboggle, $lr_{init} =0.0001$ and $\lambda=1$.
We set the batch size as $1$, $M$ as $30$.
The framework was implemented with PyTorch, using a GPU of NVIDIA RTX3090 with 24GB memory.
We set $T$ to 8 on LPBA and 5 on Mindboggle.

\begin{table}[t]
	\centering
	\caption{The numerical results of different registration methods on two datasets.}
	\label{tab1}
	\begin{tabular}{lccc|ccc}
		\toprule
		&\multicolumn{3}{c}{Mindboggle (62 ROIs)}&\multicolumn{3}{c}{LPBA (54 ROIs)}\\
		\cmidrule(l{2pt}r{2pt}){2-7}
		& DSC ($\%$) & ASSD ($mm$) & $ \% |J_{\phi}|\le0 $& DSC ($\%$) & ASSD ($mm$) & $ \% |J_{\phi}|\le0 $ \\
		\cmidrule(l{2pt}r{2pt}){1-7}
		\texttt{SyN}~\cite{AVANTS2008} & $56.7\pm1.5$ & $1.38\pm0.09$ & $\textless0.0001\%$ & $70.1\pm6.2$ & $1.72\pm0.12$ & $\textless0.002\%$  \\
		\texttt{XM}~\cite{shi2022xmorpher} & $53.6\pm1.5$ & $1.46\pm0.09$ & $\textless0.2\%$ &$66.3\pm2.0$&$1.92\pm0.15$& $\textless0.1\%$   \\
		\texttt{VM}~\cite{VoxelMorph} & $56.0\pm1.6$ & $1.49\pm0.11$ &  $\textless0.2\%$ & $68.2\pm2.3$&$1.84\pm0.17$& $\textless0.004\%$   \\
		\texttt{TM}~\cite{Chen2022a}  & $60.7\pm1.5$ & $1.35\pm0.10$ & $\textless0.2\%$ & $68.9\pm2.4$ &$1.82\pm0.18$ & $\textless0.02\%$  \\
		\texttt{DMR}~\cite{Chen2022} & $60.6\pm1.4$ & $1.34\pm0.09$ & $\textless0.09\%$ &$69.2\pm2.4$&$1.79\pm0.18$& $\textless0.2\%$   \\
		\texttt{PR++}~\cite{Dual} & $61.1\pm1.4$ & $1.34\pm0.10$ & $\textless0.08\%$ &$69.5\pm2.2$&$1.76\pm0.17$& $\textless0.06\%$   \\
		\texttt{FiRework-VM} & $60.0\pm1.6$ & $1.40\pm0.10$ & $\textless0.2\%$ & $70.4\pm2.1 $ & \boldmath{$1.71\pm0.15$} & $\textless0.02\%$ \\
		\texttt{FiRework-TM} & \boldmath{$62.8\pm1.5$}  &\boldmath{$1.30\pm0.10$} & $\textless0.2\%$ & \boldmath{$70.5\pm2.2$}  & $1.71\pm0.17$ & $\textless0.4\%$ \\
		\bottomrule
	\end{tabular}
\end{table}

\begin{figure}[t]
	\centering
	\includegraphics[width=1.0\columnwidth]{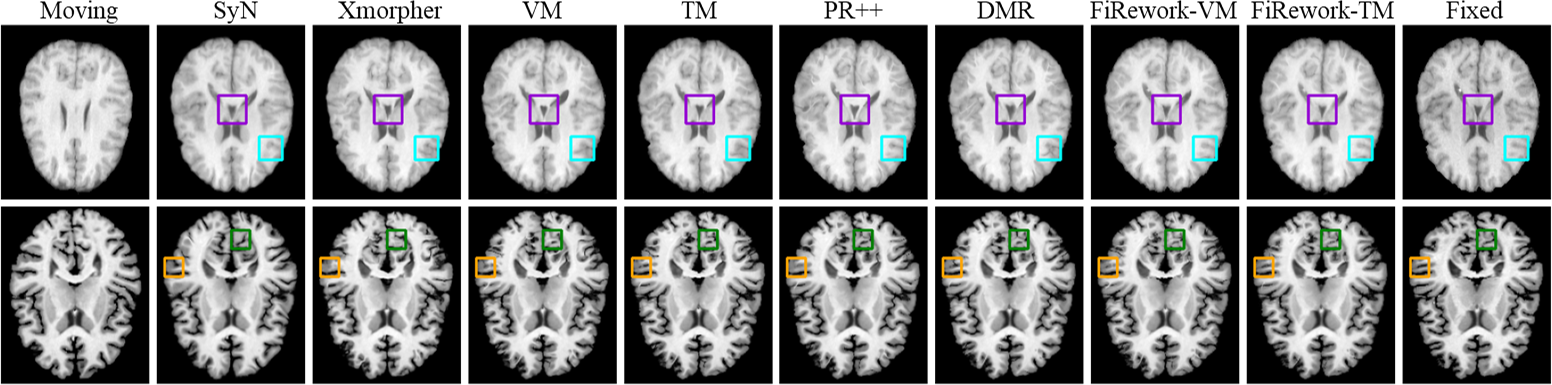}
	\caption{Registration results from different methods on LPBA (top row) and Mindboggle (bottom row).}
	\label{f:results}
\end{figure}

\subsubsection{Quantitative and Qualitative Analysis.}
The quantitative results of different methods are reported in Table~\ref{tab1}.
FiRework-TM achieved the highest DSC and lowest ASSD on Mindboggle, $62.8\%$ and $1.30$mm, respectively,
as well as the highest DSC on LPBA,
while FiRework-VM obtained the lowest ASSD on LPBA.
For the DSC results, FiRework-TM surpassed the second-best methods by $1.7\%$ and $0.4\%$ on Mindboggle and LPBA, respectively.
On the LPBA dataset, FiRework-VM and FiRework-TM achieved $2.2\%$ and $1.6\%$ improvements compared to VM and TM, respectively.
Moreover, on the Mindboggle dataset, FiRework-VM and FiRework-TM improved $4.0\%$ and $2.1\%$.
All these improvements were significant (Wilcoxon tests, $p < 0.05$).
Table~\ref{tab1} also lists the percentage of voxels with non-positive Jacobian determinant ($ \% |J_{\phi}|\le0 $).
Our method achieved acceptable performance.

\begin{figure}[t]
	\centering
	\includegraphics[width=0.9\columnwidth]{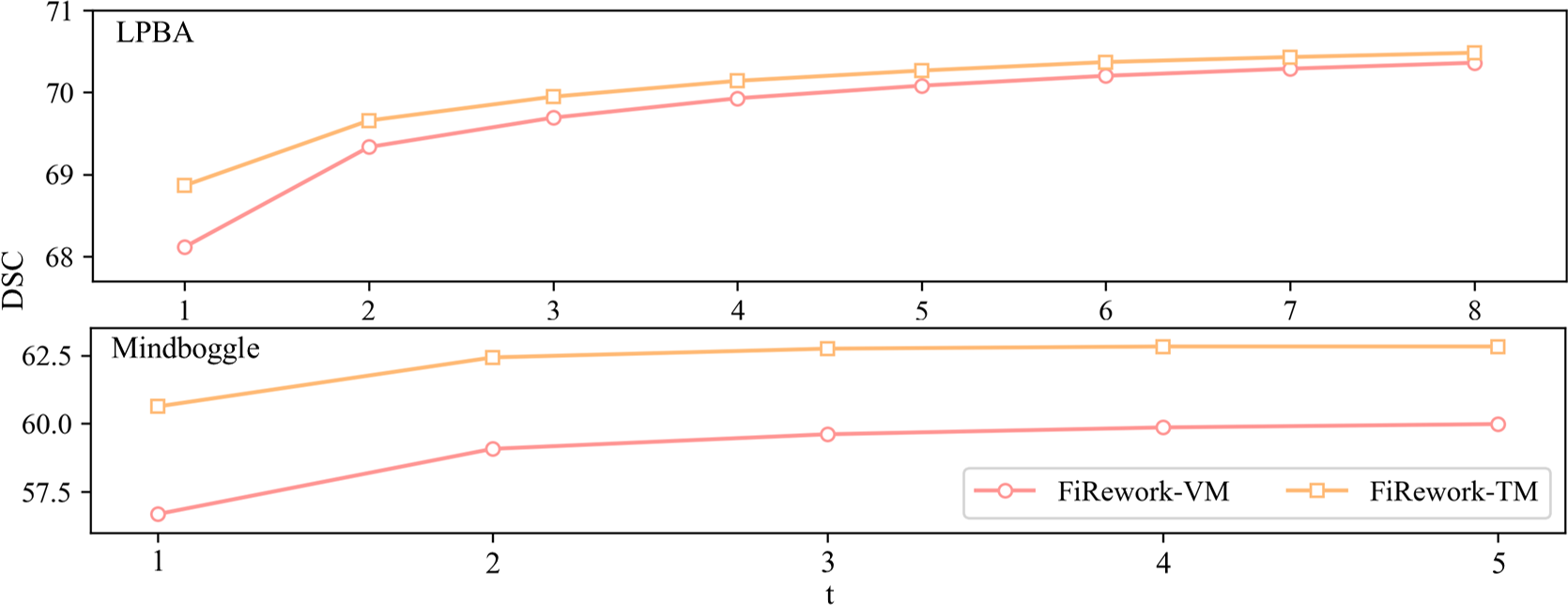}
	\caption{Illustration of the DSC values obtained for each level of continuous deformation using the methods with proposed FiRework framework on two datasets.}
	\label{f:TMVM}
\end{figure}

\begin{figure}[t]
	\centering
	\includegraphics[width=1.0\columnwidth]{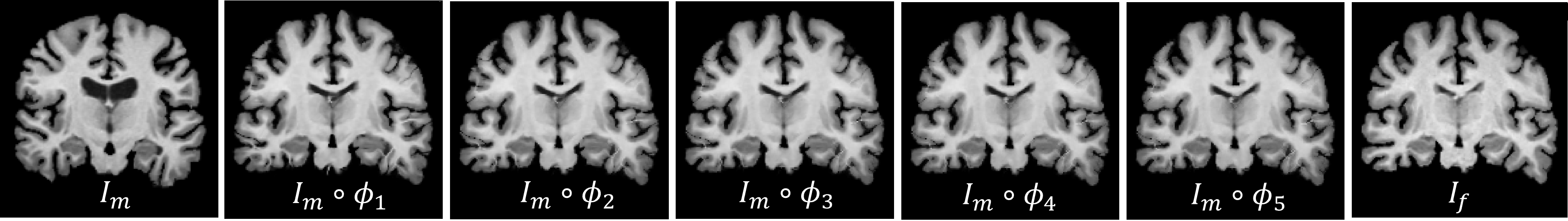}
	\caption{Visualization of the refinement process using FiRework-TM for a single image pair from the Mindboggle dataset.}
	\label{f:deformation}
\end{figure}

Fig.~\ref{f:results} shows the registration results from different methods.
Our method accurately registered corresponding structures.
Fig.~\ref{f:TMVM} illustrates the accuracy gains brought by each level of deformation in our FiRework.
Fig.~\ref{f:deformation} takes the registration of one image pair as an example to show the refinement process of FiRework-TM.
The final field $\phi_5$ accurately warped the moving image to registered with the fixed image.

It is worth noting that compared to the original VM and TM, the FiRework-VM and FiRework-TM have increased their network parameters by only approximately 0.01~MB and 0.12~MB, respectively.

\subsubsection{Analysis of Continuous Deformation Capability}

\begin{figure}[t]
	\centering
	\includegraphics[width=0.9\columnwidth]{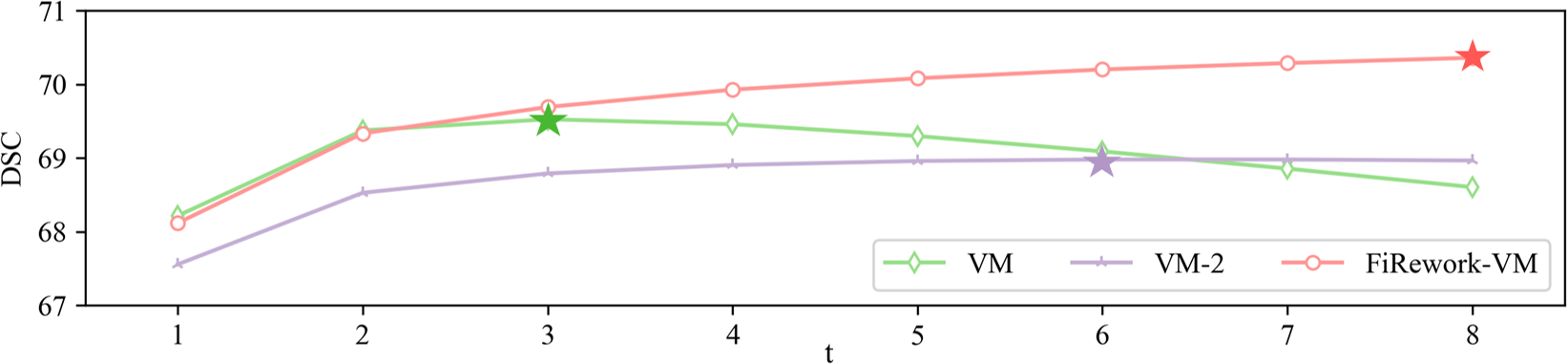}
	\caption{Comparison of continuous deformation capability on the LPBA dataset with \( T = 8 \), measured by DSC.
		VM refers to VoxelMorph (VM) without recursive training and inference (Equation~(\ref{e:continue})).
		VM-2 denotes the VoxelMorph network trained and tested using the framework depicted in Fig.~\ref{f:VM-2}, with experimental setup consistent with our method.
		FiRework-VM represents our enhanced FiRework method.
		The asterisk symbol denotes the epoch with the maximum value attained.}
	\label{f:lines}
\end{figure}

We compared our proposed framework with classical continuous deformation frameworks as shown in Fig.~\ref{f:lines}.
Choosing VM with one recursion was because its training cost is similar to that of our framework.
As shown in Fig.~\ref{f:lines}, VM reached its maximum value in the third round and then continued to decrease, mainly because the network did not know to continue cycling during learning and lacked the knowledge of how to further optimize near a local minimum point.
Furthermore, VM-2 reached its peak performance at $t = 6$ and then remained relatively stable, mainly due to cumulative errors leading to deformation getting trapped in a local minimum, and interpolation errors causing the feature points on moving image to become blurred, making it difficult to achieve a better solution using this information alone.
In contrast, FiRework-VM continuously optimized the deformation field and constantly referring to the original moving image, thus obtaining a better solution.

\section{Discussion}
We present a field refinement framework for the explicit optimization of the deformation field, where each stage takes into account the deformation field from the previous stage and predicts the errors present in the current stage's deformation field.
Furthermore, the proposed approach enables continuous optimization of the deformation field during the inference phase.
Our primary motivation of the proposed FiRework is to leverage the network modeling the deficiency of the deformation field straightforwardly.
This is different from almost all existing registration networks attempting to directly estimate the deformation field~\cite{VoxelMorph, Chen2022a, zhu2020unsupervised} or the residual subfield~\cite{cao2021edge, Dual, Lv2022, Chen2022, shi2022xmorpher}.
Compared to these existing methods, our proposed FiRework fully leverages the comprehensive information of the original image pair, the previous deformation field and its warped moving image, to refine the previous deformation field by analyzing its deficiency.
We consider this as a more efficient solution.
The efficacy of our FiRework on deformation field refinement can be observed from the quantitative and qualitative results in Table~\ref{tab1}, Fig.~\ref{f:TMVM} and Fig.~\ref{f:deformation}.

Additionally, the proposed FiRework addresses the common issue in existing multi-stage registration methods, where cumulative errors and interpolation errors often degrade the registration performance.
Our method effectively mitigates these issues, as demonstrated by its superiority shown in Fig.~\ref{f:lines}.
The experimental results in Table~\ref{tab1} further illustrate that our FiRework can significantly enhance the performance of existing registration models (e.g., VM~\cite{VoxelMorph} and TM~\cite{Chen2022a}) with minimal modifications, making it both a simple and highly effective solution.

Looking ahead, our current experiments have been limited to adapting existing networks to our FiRework.
Therefore, future work may involve designing models that are more tailored to our framework.
Additionally, our framework has not yet explored multi-resolution continuous deformation, an area we plan to explore in future research endeavors.

\section{Conclusion}
In this study, we introduce a novel field refinement framework (FiRework) for efficient enhancement of deformable registration.
The proposed FiRework addresses the issues of error accumulation and interpolation in existing continuous deformation frameworks by directly optimizing the deformation field and re-inputting the moving image at each iteration.
Our FiRework requires only one cycle during training, yet allows for multiple iterations during testing, efficiently enhancing the registration accuracy.
Experimental evaluations were conducted on two single-stage deformable registration networks using FiRework, and the results demonstrate the efficacy of our FiRework.

\section*{Acknowledgements}
This work was supported in part by the National Natural Science Foundation of China under Grants 62471306 and 62071305,
in part by the Guangdong-Hong Kong Joint Funding for Technology and Innovation under Grant 2023A0505010021,
in part by the Shenzhen Medical Research Fund under Grand D2402010,
and in part by the Guangdong Basic and Applied Basic Research Foundation under Grant 2022A1515011241.

%
%
%
 \bibliographystyle{splncs04}

\bibliography{bib2}
\end{document}